%% file: tree.tex
\documentclass[runningheads]{llncs}
\usepackage{graphicx}
\usepackage{amsmath,amssymb} 
\usepackage{color}

\input{packages}
\input{macros}

\begin{document}
\title{Seeing Tree Structure from Vibration} 

\titlerunning{Seeing Tree Structure from Vibration}
%
\author{Tianfan Xue*\inst{1} \and
Jiajun Wu*\inst{2} \and 
Zhoutong Zhang\inst{2} \and
Chengkai Zhang\inst{2} \and\\ 
Joshua B. Tenenbaum\inst{2} \and
William T. Freeman\inst{1,2}}
%
\authorrunning{T. Xue, J. Wu, Z. Zhang, C. Zhang, J. B. Tenenebaum, W. T. Freeman}
%
\footnotetext{* T. Xue and J. Wu contributed equally to this work.}
\institute{Google Research \and
MIT CSAIL}
\maketitle              

\input{text/abstract}

\input{text/intro}
\input{text/related_work}

\input{text/formulation}

\input{text/algorithm}

\input{text/evaluation}
\input{text/application}

\input{text/discussion}

\myparagraph{Acknowledgements: } 
This work is supported by NSF \#1231216, \#1212849, and \#1447476, ONR MURI N00014-16-1-2007, Shell Research, and Facebook. We thank Xiuming Zhang for helpful discussions.

%
%
%
\bibliographystyle{splncs04}
\bibliography{tree}

\appendix
\renewcommand{\thesection}{A.\arabic{section}}
\renewcommand{\thefigure}{A\arabic{figure}}
\setcounter{section}{0}
\setcounter{figure}{0}

\input{text/supp}

\end{document}

%% file: packages.tex
\usepackage{color,xcolor}
\usepackage{epsfig}
\usepackage{graphicx}
\usepackage[]{algorithm2e}

\usepackage{adjustbox}
\usepackage{array}
\usepackage{booktabs}
\usepackage{colortbl}
\usepackage{float,wrapfig}
\usepackage{hhline}
\usepackage{multirow}
\usepackage{subcaption} 
\usepackage[labelfont={bf},labelsep={period},font={small}]{caption}

\let\llncssubparagraph\subparagraph
\let\subparagraph\paragraph
\usepackage[compact]{titlesec}
\let\subparagraph\llncssubparagraph

\usepackage{amsmath,amsfonts,amssymb}
\usepackage{bm}
\usepackage{nicefrac}
\usepackage{microtype}

\usepackage{changepage}
\usepackage{extramarks}
\usepackage{fancyhdr}
\usepackage{lastpage}
\usepackage{setspace}
\usepackage{soul}
\usepackage{xspace}

\usepackage{url}

\usepackage{enumerate}

%% file: macros.tex
\newcolumntype{L}[1]{>{\raggedright\let\newline\\\arraybackslash\hspace{0pt}}m{#1}}
\newcolumntype{C}[1]{>{\centering\let\newline\\\arraybackslash\hspace{0pt}}m{#1}}
\newcolumntype{R}[1]{>{\raggedleft\let\newline\\\arraybackslash\hspace{0pt}}m{#1}}

\newcommand{\sect}[1]{Section~\ref{#1}}
\newcommand{\sects}[1]{Sections~\ref{#1}}
\newcommand{\eqn}[1]{Equation~\ref{#1}}
\newcommand{\eqns}[1]{Equations~\ref{#1}}
\newcommand{\fig}[1]{Figure~\ref{#1}}

\newcommand{\tbl}[1]{Table~\ref{#1}}
\newcommand{\tbls}[1]{Tables~\ref{#1}}

\newcommand{\ignore}[1]{}

\makeatletter
\DeclareRobustCommand\onedot{\futurelet\@let@token\@onedot}
\def\@onedot{\ifx\@let@token.\else.\null\fi\xspace}

\def\eg{\emph{e.g}\onedot} 
\def\ie{\emph{i.e}\onedot}

\def\etal{\emph{et al}\onedot}

\makeatother

\definecolor{MyDarkBlue}{rgb}{0,0.08,1}
\definecolor{MyDarkGreen}{rgb}{0.02,0.6,0.02}
\definecolor{MyDarkRed}{rgb}{0.6,0.02,0.02}
\definecolor{MyDarkOrange}{rgb}{0.40,0.2,0.02}
\definecolor{MyPurple}{RGB}{111,0,255}
\definecolor{MyRed}{rgb}{1.0,0.0,0.0}
\definecolor{MyGold}{rgb}{0.75,0.6,0.12}
\definecolor{MyDarkgray}{rgb}{0.66, 0.66, 0.66}

\def\defeq{\stackrel{def}{=}}

\newcommand{\fa}[1]{\boldsymbol{a}_{#1}(\htheta,\dhtheta,\ddhtheta)}
\newcommand{\fr}[1]{\boldsymbol{r}_{#1}(\htheta,\dhtheta,\ddhtheta)}

\newtheorem{prop}{Property}

\def\x{\bm{x}}

\def\hy{\hat{y}}
\def\y{\bm{y}}
\def\ya{y^{1}}
\def\yb{y^{2}}

\def\ddy{\ddot{\y}}
\def\n{\bm{n}}

\def\ri{\bm{r}_i}
\def\rl{\bm{r}_l}
\def\rc{\bm{r}_c}

\def\bhtheta{{\bm{\theta}}}
\def\htheta{{\theta}}
\def\dhtheta{\dot{{\theta}}}
\def\ddhtheta{\ddot{{\theta}}}
\def\ai{\bm{a}_i}

\def\aio{\bm{a}_{i_o}}
\def\apo{\bm{a}_{p_o}}
\def\g{\bm{g}}
\def\h{\bm{h}}
\def\R{\mathbb{R}}

\newcommand{\myfirstparagraph}[1]{\noindent{\bf #1}}
\newcommand{\myparagraph}[1]{\vspace{3pt}\noindent{\bf #1}}

%% file: text/abstract.tex
\begin{abstract}

Humans recognize object structure from both their appearance and motion; often, motion helps to resolve ambiguities in object structure that arise when we observe object appearance only. There are particular scenarios, however, where neither appearance nor spatial-temporal motion signals are informative: occluding twigs may look connected and have almost identical movements, though they belong to different, possibly disconnected branches. We propose to tackle this problem through spectrum analysis of motion signals, because vibrations of disconnected branches, though visually similar, often have distinctive natural frequencies. We propose a novel formulation of tree structure based on a physics-based link model, and validate its effectiveness by theoretical analysis, numerical simulation, and empirical experiments. With this formulation, we use nonparametric Bayesian inference to reconstruct tree structure from both spectral vibration signals and appearance cues. Our model performs well in recognizing hierarchical tree structure from real-world videos of trees and vessels.

\keywords{Vibration \and Tree structure \and Hierarchical Bayesian model}

\end{abstract}

%% file: text/intro.tex
\input{figText/teaser.tex}

\section{Introduction}
\label{sec:intro}

In visual perception, motion information often helps to resolve appearance ambiguities. Animals may conceal themselves with camouflaged clothing, but they are unlikely to match their motion with that in the background, such as foliage waving in the breeze~\cite{davies2012perception}. In medical imaging, it might be hard to separate blood vessels (or fibers) purely from their appearance, but the distinction becomes clear once the vessels start to vibrate. Extensive studies in cognitive science also suggest that humans, including young children, recognize objects from both appearance and motion cues~\cite{spelke1992origins}.

Computer vision researchers have combined motion and appearance information to solve a range of tasks~\cite{bascle1996motion,pathak2017learning}. Bouman~\etal proposed to estimate physical object properties based on their appearance and vibration~\cite{bouman2013estimating}. Wang~\etal proposed a layered motion representation~\cite{wang1993layered}, which has been widely employed in object segmentation and structural prediction~\cite{jepson2002layered,sun2014local}. 

In this paper, we focus on tree structure estimation. This problem is even more challenging, as both motion and appearance cues can fail to discriminate pixels of disjoint branches.  We show an example in \fig{fig:teaser}. The three points $\{P_i\}$ in \fig{fig:teaser} are on two occluding branches. There are two plausible explanations: either $P_1$ and $P_2$, or $P_1$ and $P_3$ may be on the same branch. Due to self-occlusion, it is hard to infer the underlying connection just from their appearance. It is also challenging to resolve this ambiguity using only temporal motion information: the movement of these three nodes are dominated by the vibration of the root branch, so they share almost the same trajectories (\fig{fig:teaser}c).

We propose to incorporate spectral analysis to deal with this problem. This is inspired by our observation that pixels of different branches often have distinctive modes in their spectra of frequency responses, despite their similar spatial trajectories. As shown in \fig{fig:teaser}d, $P_3$ has distinct amplitude at certain frequencies compared with $P_1$ and $P_2$; intuitively and theoretically (discussed in \sect{sec:formulation}), $P_3$ is more likely to be on a separate branch. 

Our formulation of tree vibration builds upon and extends a physics-based link model from the field of botany~\cite{murphy2012physics}. Here, we deduce a key property of tree structure: each branch is a linear time-invariant (LTI) system with respect to the vibration of root. With this property, we can infer the natural frequencies of each sub-branch in a tree from its frequency response, and group nodes based on the inferred natural frequencies. We also provide justifications of this property through theoretical analysis, numerical simulation, and empirical experiments.

Based on our tree formulation, we develop a hierarchical grouping algorithm to infer tree structure, using both spectral motion signals and appearance cues. As each node in a tree may connect to an indefinite number of children, our inference algorithm employs nonparametric Bayesian methods.

For evaluation, we collect videos of both artificial and real-world tree-structured objects. We demonstrate that our algorithm works well in recognizing tree structure, using both appearance cues and spectra of vibration. We compare our algorithm with baselines that use spatial motion signals; we also conduct ablation studies to reveal how each component contributes to the algorithm's final performance. Our model has wide applications, as tree structure exists extensively in real life. Here we show two of them: seeing shape from shadow, and connecting blood vessels from retinal videos.

Our contributions are three-fold. Our main contribution is to show that tiny, barely visible object motion can reveal object structure. Our model can resolve the ambiguity in tree structure estimation using spectral information. Second, we propose a novel, physics-based tree formulation, with which we may estimate the natural frequencies of each sub-branch. Third, we design a hierarchical inference algorithm, using nonparametric Bayesian methods to predict tree structure. Our algorithm achieves good performance on real-world videos. 

%% file: figText/teaser.tex
\begin{figure}[t]
    \centering
    \includegraphics[width=\textwidth]{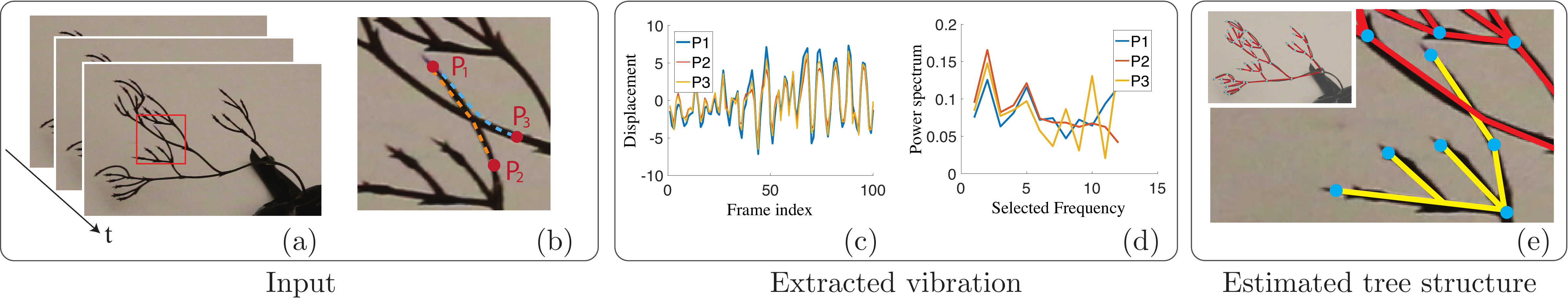}
    \vspace{-15pt}
    \caption{We want to infer the hierarchical structure of the tree in video (a). Inference based on a single frame has inherent ambiguities: figure (b) shows an example, where it is hard to tell from appearance whether point $P_1$ is connected to $P_2$ (orange curve) or to $P_3$ (blue curve). Time domain motion signals do not help much, as these branches have almost identical movements (c). We observe that the difference is significant in the frequency domain (d), from which we can see $P_1$ is more likely to connect to $P_2$ due to their similar spectra. We therefore develop an algorithm that infers tree structure based on both vibration spectra and appearance cues. The results are shown in (e).}
    \vspace{-20pt}
    \label{fig:teaser}
\end{figure}

%% file: text/related_work.tex
\section{Related Work}

\myfirstparagraph{Motion for Structured Prediction. } 
Researchers in computer vision have been using motion signals for various tasks~\cite{bascle1996motion,pathak2017learning,sun2012layered,xue2015computational}. For structured prediction in particular, the layered motion representations~\cite{wang1993layered} have been studied and applied extensively~\cite{jepson2002layered,sun2014local}. These papers model motion signals in the temporal domain; they are not for scenarios where objects may only have subtle motion differences.

Regarding spectral analysis of motion, the pioneer work of Fleet and Jepson~\cite{fleet1990computation} discussed how phase signals could help to estimate object velocity. Gautama and Van~\cite{gautama2002phase} extended the work, proposing a phase-based approach for optical flow estimation. Zhou~\etal~\cite{zhou2010phase} also discussed how phase information helps recognizing object motion. Recently, there have also been a number of works on visualizing and magnifying subtle motion signals from video~\cite{wu2012eulerian,davis2015visual}, and Rubinstein~\etal did a thorough review in~\cite{rubinstein2013analysis}.

The problem of tree structure estimation has been widely studied in computer vision, especially in medical imaging~\cite{fraz2012blood,turetken2011automated,turetken2016reconstructing,wang2011novel}, mostly from a static image. In this paper, we explore how motion signals in a video could help in structured prediction, in addition to appearance cues. Though we currently employ a simple and intuitive appearance model, it is straightforward to incorporate more sophisticated appearance models into our approach.

\myparagraph{Modeling Tree Vibration. }
Tree vibration is an important research area in the field of botany~\cite{james2014tree,moore2004natural}. Moore and Maguire~\cite{moore2004natural} reviewed the concepts and dynamic studies by examining the natural frequencies and damping ratios of trees in winds. Recently, James \etal~\cite{james2014tree} reviewed tree bio-mechanics studies using dynamic methods of analysis.

Our formulation of tree vibration is based on the lumped-mass procedure. Related literature include spring-mass-damper models for trees as a single mass point~\cite{miller2005structural}, or as a complex system of coupled masses that represent the trunk and branches~\cite{james2006mechanical,murphy2012physics}. Our formulation also considers a tree as a system of coupled masses, but different from Murphy~\etal\cite{murphy2012physics} which studied only one-layer structure, we explore hierarchical tree structure of multiple layers.

\myparagraph{Bayesian Theory of Perception. }
Researchers have developed Bayesian theories for human visual perception in general~\cite{knill1996perception,lee2003hierarchical,moreno2011bayesian}, and for object motion perception in  particular~\cite{braddick1993segmentation,weiss1998slow}. Our inference algorithm draws inspirations from the recent hierarchical Bayesian model for object motion from Gershman~\etal~\cite{gershman2016discovering}, which employs the nested Chinese restaurant process (nCRP)~\cite{blei2010nested} as a prior of object structure.

%% file: text/formulation.tex
\section{Formulation}
\label{sec:formulation}

\input{figText/formulation}

We here present our formulation that recovers tree structure from the temporal complex spectra of vertices. We start by introducing a physics-based, hierarchical link model, representing a tree as a set of beams with certain mass and stiffness (\fig{fig:tree}). Using this model, we derive a set of ordinary differential equations (ODEs) of node vibrations (\sect{sec:ODE}) and prove an important property (\sect{sec:properties}): each sub-branch of a tree is a linear time-invariant system under certain assumptions. A Bayesian inference algorithm exploits the property for structure estimation (\sects{sec:recognition} and \ref{sec:inference}).


\subsection{A Physics-Based Link Model}
\label{sec:problem}

We use a rigid link model to describe the vibration of a tree, as shown in \fig{fig:tree}. In this model, each branch $i$ of the tree is modeled as a rigid beam with a certain mass $m_i$ and length $l_i$. Under the uniform mass assumption, the center of mass of a branch is at $\frac{l_i}{2}$. Each branch connects to its parent through a torsional spring with stiffness $k_i$. Our model relates to the simpler, one-layer physical model from Murphy~\etal~\cite{murphy2012physics}, where they attempted to compute the mass and stiffness of all the beams. We observe this to be impractical in real data given the presence of noise and occlusion. Instead, we derive a set of non-linear ordinary derivative equations (ODEs) that describe the relationship between the vibration of a tree and its structure and physical properties.

We describe the vibration of a tree by the deviation angles $\{\htheta_i\}$\xspace of branches. As shown in~\fig{fig:beam}, let $\hat\htheta_i$ be the directional angle from vertical line to a branch  when the tree is static (no external forces except gravity), and let $\htheta_i$ be the deviation angle from its static location when the tree is vibrating ($\theta_i$ changes over time). To derive the governing equations for $\htheta_i$, we start by applying the Newton's law to each branch $i$, which gives\footnote{In this chapter, we use a lower-case letter $a$ to denote a scalar, a bold lower-case letter $\bm{a}$ to denote a vector, and a capital letter $A$ to denote a matrix. We denote the matrix product as $A\bm{b}$, where $A \in \R^{n \times m}$ and $\bm{b} \in \R^m$.}
\vspace{-5pt}
\begin{equation}
    m\ai = -\ri + \sum_{c \in C_i} \rc + m\g, \label{eqn:acc_vec}
\vspace{-5pt}
\end{equation}
where $\rc \in \R^2$ is the force exerted by branch $c$ on its parent, $C_i$ is the set of children of branch $i$, and $\g$ is the acceleration due to gravity. The negative sign before $\ri$ is due to our definition and Newton's third law. Branch $i$'s acceleration $\ai \in \R^2$ is defined as the acceleration of the branch's center of mass. 

In addition, we have the rotation equation,
\vspace{-5pt}
\begin{equation}
   I_i \dot{\omega}_i = -k_i \htheta_i + \sum_{c\in C_i} k_c \htheta_c + \ri \times \x_i + \sum_{c\in C_i}\rc \times \x_i,\label{eqn:rot_vec}
\vspace{-5pt}
\end{equation}
where $I_i$ is branch $i$'s moment of inertia when it rotates around its center, $\dot{\omega}_i$ is its angular acceleration, $\theta_c$ is branch $c$'s deviation angle, $\x_i$ is its movement, and $k_i$ is the stiffness of the torsional spring it connects to. Also, the branch acceleration $\ai$ relates to the acceleration of its endpoint $\aio$ via
\vspace{-5pt}
\begin{equation}
    \ai = \aio + \dot{\omega}_i \times \x_i + \omega_i \times (\omega_i \times \x_i), \label{eqn:acc_constraint_vec}
\vspace{-5pt}
\end{equation}
where $\aio \in \R^2$ is the acceleration of the junction point.

Therefore, the angular velocity and angular acceleration of branch $i$ are
\vspace{-5pt}
\begin{equation}
    \omega_i = \dhtheta_i + \sum_{p\in P_i}\dhtheta_p \quad\text{and}\quad 
    \dot{\omega}_i = \ddhtheta_i + \sum_{p\in P_i}\ddhtheta_p, \label{eqn:agacc}
\vspace{-5pt}
\end{equation}
where $P_i$ is the set of ancestors of branch $i$. These equations do not include fictitious forces. All quantities are global values under the reference frame.

At last, replacing the branch acceleration ($\ai$ and $\aio$) and angular acceleration $\dot{\omega}_i$ in \eqns{eqn:acc_vec} and \ref{eqn:rot_vec} using \eqns{eqn:acc_constraint_vec} and \ref{eqn:agacc}, and eliminating forces between branches $r_i$, we get the ODE with respect to all deviation angles $\{\theta_i\}$,
\vspace{-5pt}
\begin{equation}
\small
    I_i f_{i}(\ddhtheta)  = -k_i \htheta_i + \sum_{c\in C_i} k_c \htheta_c + \fr{i} \times \x_i + \sum_{c\in C_i}\fr{c} \times \x_i, \label{eqn:nl_ode}
\vspace{-5pt}
\end{equation}
where $\fr{i}$ is a vector functions of $\htheta$, $\dhtheta$, and $\ddhtheta$. Please see our supplementary material for its definition in detail.


\subsection{ODE of Node Vibration}
\label{sec:ODE}

The ODE (\eqn{eqn:nl_ode}) is highly nonlinear due to sinusoidal and quadratic terms. To solve it, we first linearize the equation around its stable solution. We assume that the deviation angle $\htheta_i$ of each branch $i$ is small and ignore all $O(\htheta_i^2)$ terms. Under this assumption, the quadratic term of angular velocity  $O(\dhtheta^2)$ can also be ignored, because according to the conservation of energy, the potential energy $\frac{1}{2}k\htheta^2$ of a branch is on the same scale of its kinetic energy $\frac{1}{2}I_i\dhtheta^2$. 

We can now derive a fully linear system under the above assumption as
\vspace{-5pt}
\begin{equation}
    M\ddot{\bhtheta} + K\bhtheta = \bm{0},
    \label{eqn:theta_matrix_form}
\vspace{-5pt}
\end{equation}
where $M$ and $K$ are two matrices depending on the structure of a tree and its physical properties, including the moment of inertia ($I$), mass ($m$), and stiffness ($k$) of all branches.

In practice, from an input video, it is easier to measure the 2D shift of each node, rather than the rotation of each branch. To derive the ODE of 2D shifts of all nodes from \eqn{eqn:theta_matrix_form}, we denote node $i$'s 2D location in a stable tree as $\hat{\bm{y}}_i$, and the 2D shifts from its stable location as $\bm y_i$. We have
\vspace{-5pt}
\begin{equation}
\bm y_i + \bm \hy_i = \sum_{j\in P_i} l_j \n(\theta_j+\hat \theta_j), \label{eqn:x_theta_relation}
\vspace{-5pt}
\end{equation}
where $\n(\theta) = (\cos\theta, \sin\theta)$ and $l_j$ is the length of branch $j$ (recall that $P_i$ is the set of ancestors of branch $i$). Let $\y$ be the concatenation of 2D shifts of all the nodes. Plugging \eqn{eqn:x_theta_relation} to \eqn{eqn:theta_matrix_form}, we have
\vspace{-5pt}
\begin{equation}
N \ddy + L \y = \bm{0},
\vspace{-5pt}
\label{eqn:ODE_y}
\end{equation}
where $N$ and $L$ are matrices depending on $M$, $K$, $l_j$, and $\theta_j$. The constant term must be zero, as $\y = \ddy = \bm{0}$ when the tree is stable. Please see our supplementary material for a detailed derivation.


\subsection{Inferring Modes of Each Sub-branch}
\label{sec:properties}

Based on the second order ODE, we can infer the modes of each sub-branch and use them to group nodes into branches using the following property.

\begin{prop}[\textbf{Each sub-branch is a LTI-system}]
\vspace{-8pt}
Imagine a branch undergoes a forced vibration. Let $y_{root}^i(t)$ and $y_{leaf}^i(t)$ be the displacements of the root and one of its leaf node respectively at time $t$ ($i=1,2$). Then, if the displacement of the root is $\alpha_1 \cdot \ya_{root}(t) + \alpha_2 \cdot \yb_{root}(t)$, where $\alpha_1,\alpha_2 \in \R$, the vibration of the leaf is
$\alpha_1 \cdot \ya_{leaf}(t) + \alpha_2 \cdot \yb_{leaf}(t).$    
\label{prop:1}
\vspace{-8pt}
\end{prop}

This is a corollary of \eqn{eqn:ODE_y}, which shows that the displacement of a node satisfies a linear, second order ODE. The system is also time-invariant, as all matrices in \eqn{eqn:ODE_y} do not change in time. 

The key observation of our work is that we can infer the mode of free vibration of each sub-branch as if that sub-branch is disconnected from the rest of the tree, as suggested by Property \ref{prop:1}. Let $S$ be a set of nodes in a sub-branch; let $Y_i(\eta)$ be the temporal spectrum of the displacement of the $i$-th node in that branch ($i \in S$), where $\eta$ is the frequency index; let $Y_{root}$ be the temporal spectrum of the root displacement. Because each sub-branch is a LTI-system, the frequency response of the sub-branch is
\vspace{-5pt}
\begin{equation}
\overline{Y}_i(\eta) = \frac{Y_i(\eta)}{Y_{root}(\eta)}, \quad \forall \eta.
\vspace{-5pt}
\label{eqn:divide_spectrum}
\end{equation}
It is well known that when there is no damping, the natural frequencies of an oscillating system coincide with its resonance frequency~\cite[Chapter~4]{french1971vibrations}. In our case, this suggests that the natural frequencies of a sub-branch are the same as the modes of the frequency response of that branch\footnote{In the presence of small damping, the difference between the modes of frequency response and the modes of free vibration is also small.}.

As an illustration, \fig{fig:LTI}a shows a tree with two sub-branches ($Y_{2-4}$ and $Y_{5-7}$). All nodes have similar power spectra as their vibrations are dominated by the vibration of the root ($Y_1$). To distinguish the spectra of the two sub-branches, we calculate the frequency response of each node, \ie, the ratio between the spectrum of the root and the spectrum of each branch. As shown in \fig{fig:LTI}b, there is a clear difference between the frequency responses of two branches. The modes of each frequency response also match the modes of free vibrations of each sub-branch, as if they are detached from the root (see \fig{fig:LTI}c and d).

We can then group nodes into different sub-branches based on their spectrum response, because the natural frequencies of each sub-branch depend on its inherent physical properties like mass and stiffness. In practice, the modes of frequency responses are not a robust measure in the presence of noise and damping. Therefore, we group nodes based on their the normalized power spectra and phases instead, with the help of the appearance information described in~\sect{sec:recognition}.

\input{figText/LTI}

%% file: figText/formulation.tex
\begin{figure}[t]
\centering
\begin{subfigure}[b]{0.23\linewidth}
\includegraphics[width=\linewidth]{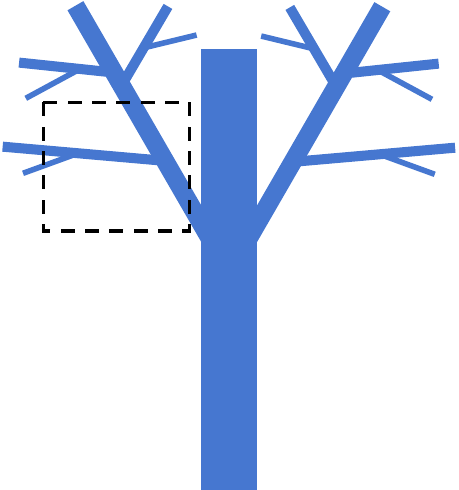}
\caption{}
\label{fig:tree}
\end{subfigure}
\qquad
\begin{subfigure}[b]{0.35\linewidth}
\includegraphics[width=\linewidth]{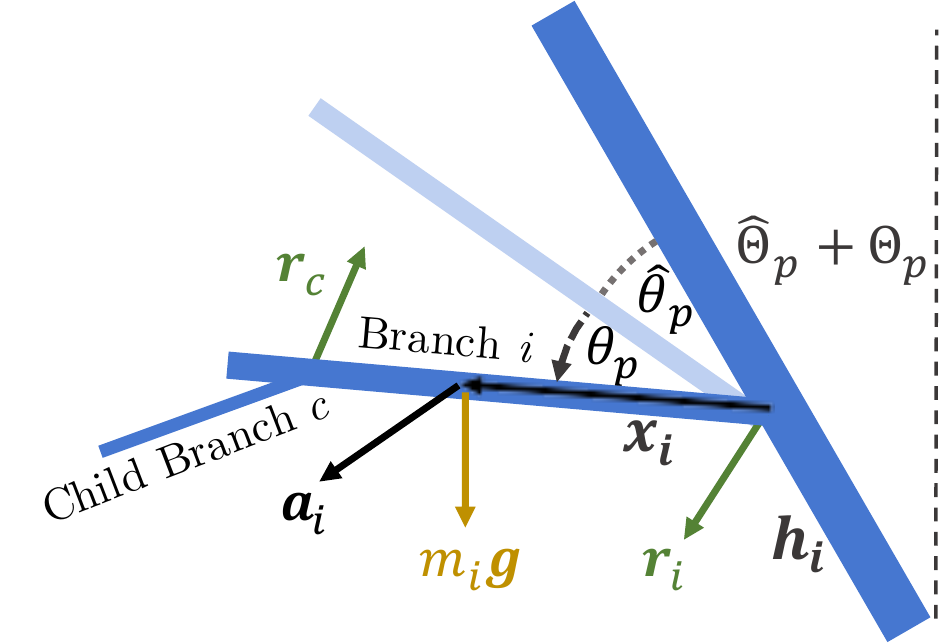}
\caption{}
\label{fig:beam}
\end{subfigure}
\vspace{-10pt}
\caption{(a) Hierarchical beam structure. (b) Force analysis for one of the branches (the one marked by dashed rectangle in (a)).}
\label{fig:h_beam}
\vspace{-20pt}
\end{figure}

%% file: figText/LTI.tex
\begin{figure}[t]
\centering
\includegraphics[width=.7\linewidth]{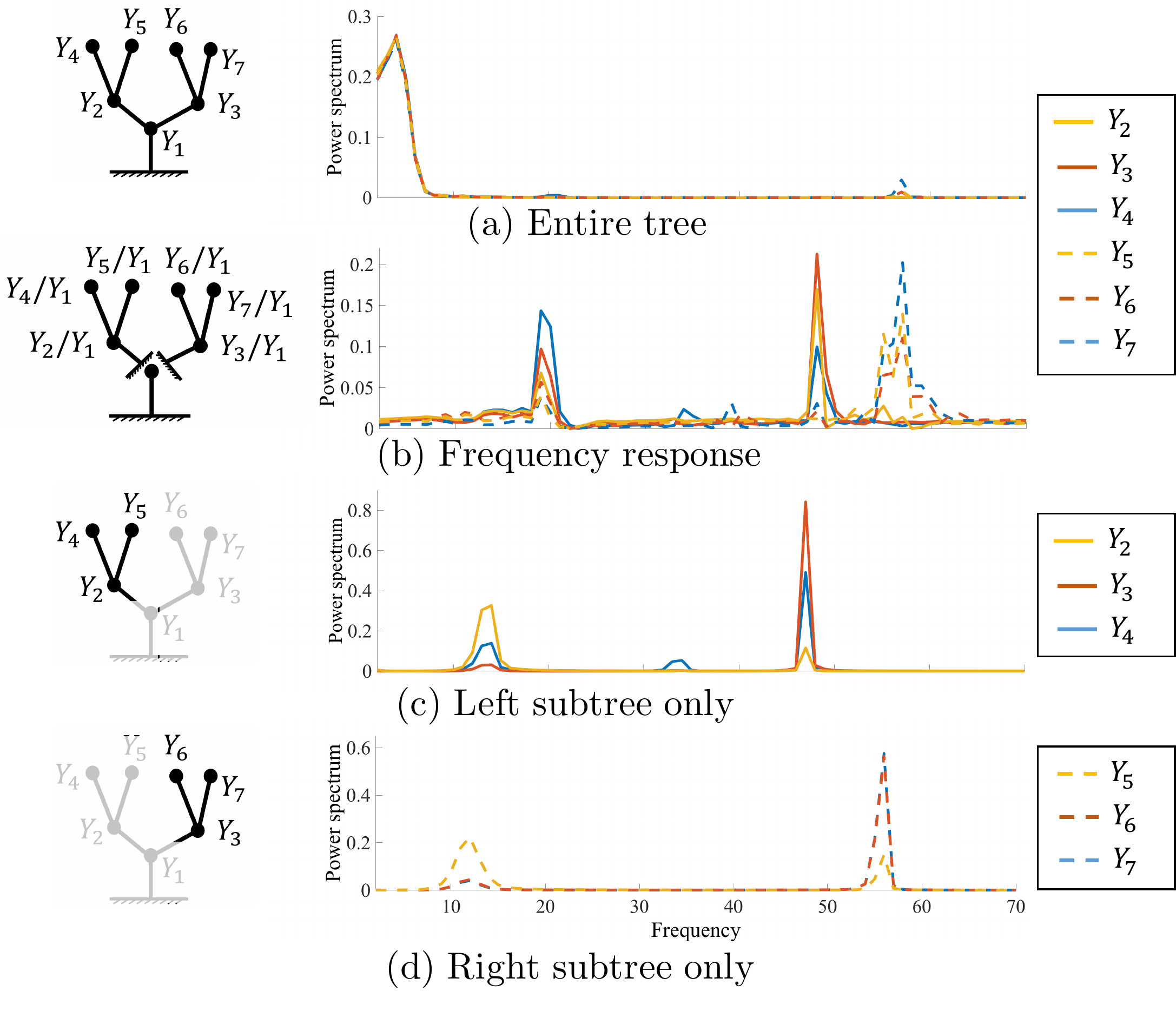}
\vspace{-10pt}
\caption{Spectrum analysis on a synthetic tree. Directly calculating the power spectrum of the vibration of each nodes does not help to infer the tree structure, as all the nodes have similar power spectrum (a). By dividing the spectrum of each node by the spectrum of the root node, we obtain the frequency response of each node. We now clearly see the difference between the two subtrees (b). The modes of each frequency response also match the modes of the free vibration of each subtree (c) and (d).}
\label{fig:LTI}
\vspace{-22pt}
\end{figure}

%% file: text/algorithm.tex
\section{Algorithm}
\label{sec:algorithm}

\input{figText/pipeline}

We now introduce our structure estimation algorithm based on the tree formulation. Our algorithm has two major components: a recognition module that extracts motion and appearance cues from visual input, and an inference module that predicts tree structure. 

\subsection{Extracting Motion and Appearance Cues}
\label{sec:recognition}

We use an bottom-up recognition algorithm to obtain motion and appearance cues from input videos (\fig{fig:pipeline}a) with a given set of interest points (\fig{fig:pipeline}b).

\myparagraph{Motion. } Given an input video, we first manually label all nodes in the first frame and then track them over time using optical flow. There are many tracking algorithms that can extract trajectories of sparse keypoints~\cite{hare2016struck,henriques2015high,lucas1981iterative,Rubinstein12Towards}, but we choose to calculate the dense motion field for two reasons. First, most of vibrations are small, and optical flow is known to perform well on capturing the small motion with subpixel accuracy. Second, sparse tracking algorithms, like the KLT tracker~\cite{lucas1981iterative}, might suffer the aperture problem, as most of branches only contain one-dimensional local structure. On the other hand, dense optical flow algorithms aggregate the information from other locations, so it would be more robust to the aperture problem.

Specifically, we first compute a dense flow field from the first frame to one of the frame $t$ in the sequence~\cite{liu2009beyond}. We then get trajectory of each node in the sequences from dense motion fields through interpolation. We further apply Fourier transform to the trajectory of each node independently to get its complex spectrum $Y$ (\fig{fig:pipeline}-II), and extract its modes from the fifth order spectral envelope~\cite{furoh2013detection}. We use the normalized amplitude  (\fig{fig:pipeline}c) and phase (\fig{fig:pipeline}d) of these modes for inference, as discussed in \sect{sec:inference}.

\myparagraph{Appearance. } We use an over-complete connectivity matrix as our appearance cues. As shown in \fig{fig:pipeline}-III, we compute the matrix via the following steps: obtaining a contour map, computing the closure of each interest point, flood-filling the contour map from all closures, and adding edges to junctions.

Given the first frame from an input video, we first use Canny edge detector~\cite{canny1986computational} with threshold 0.5 to obtain an initial contour map (\fig{fig:pipeline}e). Then, for each interest point $i$, we consider all contour pixels $S_i$ whose distance to $i$ is no larger than $r_i$. We search for the minimum $r_i$, such that if we connect $i$ to all pixels in $S_i$, the angle between each two adjacent lines is no larger than $30^{\circ}$. We call $S_i$ the closure for point $i$ (\fig{fig:pipeline}f). 

We then apply a shortest-path algorithm to obtain the connectivity map of all interest points. Our algorithm is a variant of the Dijkstra's algorithm~\cite{dijkstra1959note}, where there is a hypothetical starting point connecting to pixels in the union of all closures with cost 0. The cost between two 8-way adjacent pixels is 0, if they are both on the contour map, or 1 otherwise. The algorithm is then in essence expanding all closures simultaneously. When it finishes, we connect two keypoints if their corresponding closures are adjacent after expansion (\fig{fig:pipeline}g). To balance the expansion rate of each closure, we use a tuple $(c_i, d_i)$ as the entry for any pixel $i$ in the priority queue, where the primary key $c_i$ is the traditional term for the distance on the graph from $i$ to the origin, and the secondary key $d_i$ is the Chebyshev ($L_\infty$) distance between $i$ to the center of its closure.

Finally, an observed junction in a 2D image may be an actual tree fork, or may be just two disconnected, overlapping branches. To deal with the case, for all points that have 4 or more neighbors, we add an edge between each pair of its neighbors whose angle is no smaller than $135^{\circ}$. This leads to an over-complete connectivity matrix $E$ (\fig{fig:pipeline}h), which we use as our appearance cues. 
\input{figText/algorithm} 
\input{figText/cluster}

\subsection{Inference}
\label{sec:inference}

\myfirstparagraph{Overview with a Toy Example. } We start with a high-level overview of our hierarchical inference algorithm along with a toy tree with three levels of hierarchy (\fig{fig:h_cluster}). As shown in Algorithm~\ref{alg:1}, given the root, our algorithm first computes the free vibration of the rest of nodes (Step I), groups them into several clusters (Step II), and then recursively finds tree structure for each cluster (Step III).

In this toy tree with $v_1$ as the root, the algorithm groups the other nodes into two clusters: $(v_2, v_4, v_5)$ and $(v_3, v_6, v_7, \dots, v_{11})$, as shown in \fig{fig:cluster_b}. For each subtree, the algorithm recursively applies itself for finer-level tree structure. Here in the right branch, we get two level-$2$ subtrees $(v_6, v_8, v_9)$ and $(v_7, v_{10}, v_{11})$.

\myparagraph{Step I: Computing Free Vibration. }
We first compute the vibration of each node given the root. Based on \eqn{eqn:divide_spectrum}, we divide the complex spectrum of each leaf node by the complex spectrum of the root. Note that under certain frequency, the complex spectrum of the root might be close to zero. Therefore, a direct division might magnify the noise. To deal with this, we calculate the spectrum of each node $i$ after removing the root $r$ via
$Y_i \cdot Y_{r}^*/\left(|Y_{r}|^2 + \epsilon^2\right)$,
where $Y_{r}^*$ is the complex conjugate of $Y_{r}$, and $\epsilon$ controls the noise level. This is similar to the Weinner filter~\cite{wiener1949extrapolation}. When $\epsilon = 0$, We have the normal division as
\vspace{-5pt}
\begin{equation}
\frac{Y_i \cdot Y_{r}^*}{|Y_{r}|^2} = \frac{Y_i \cdot Y_{r}^*}{Y_{r} \cdot Y_{r}^*} = \frac{Y_i}{Y_{r}}. 
\vspace{-5pt}
\end{equation}

\myparagraph{Step II: Grouping Nodes.}
We group nodes into clusters $\{S_j\}$ under the assumption that nodes in each cluster share similar vibration patterns (complex frequencies) and appearance cues.
Each node has an unknown number of children, we use a Chinese Restaurant Process (CRP) prior~\cite{blei2010nested} over the tree structure. Let $z_i$ be the index of cluster that node $i$ is assigned to, and let $Z=\{z_i\}$ be the assignment of all nodes. The joint probability of assignment is
\vspace{-5pt}
\begin{equation}
 P(Z | E, Y) \propto \; P_\text{CRP}(Z) \cdot P_m(Y | Z) \cdot P_a(E | Z), \label{eq:joint_prob}
\vspace{-5pt}
\end{equation}
where $P_\text{CRP}(\cdot)$ is the CRP prior, $P_m(\cdot)$ is the likelihood based on motion, and $P_a(\cdot)$ is the likelihood based on appearance.

\vspace{3pt}\noindent
\emph{Motion term:} we use two statistics of the spectrum: the normalized amplitude ${Y}^n_i= |Y_i| / \|Y_i\|_2$ and the phase ${Y}^p_i= \text{angle}(Y_i)$. Our motion term is
\vspace{-5pt}
\begin{equation}
\small
\log P_m(Y | Z) = \sum_i -\sigma_n^{-2} \|{Y}^n_i - {C}^n_{z_i}\|_2^2 - \sigma_p^{-2} \|{Y}^p_i - {C}^p_{z_i}\|_2^2.
\vspace{-10pt}
\end{equation}
${C}^n_{k}$ and ${C}^p_{k}$ are the mean normalized amplitudes and phases of nodes in cluster $k$.

\vspace{3pt}\noindent
\emph{Appearance term:} nodes in the same sub-branch are expected to be connected to each other and to the root. To this end, we define the appearance term as
\vspace{-5pt}
\begin{equation}
\small
\log P_a(E | Z) = \sum_{z_i = z_j} \alpha\cdot\bm{1}(i, j|Z,E) + \sum_{i} \beta\cdot\bm{1}(i, r|Z, E),
\vspace{-10pt}
\end{equation}
where $\bm{1}(i, j)$ is the indicator function of whether there exists a path between nodes $i$ and $j$ given the current assignment $Z$ and the estimated connectivity matrix $E$ (see \sect{sec:recognition}). Given the joint probability in \eqn{eq:joint_prob}, we run Gibbs sampling~\cite{geman1984stochastic} for 20 iterations over each assignment $z_i$. 

\myparagraph{Step III: Recursion. }
As shown in the toy example (\fig{fig:h_cluster}), for each cluster $S_j$, our algorithm selects the node closest to the root $r$ in the Euclidean space as the subroot $r_j$. It then infers subtree structure for $S_j$ recursively. The whole inference algorithm takes 3--5 seconds for a tree of 50 vertices on a Desktop CPU.

%% file: figText/pipeline.tex
\begin{figure*}[t]
\centering
\includegraphics[width=\linewidth]{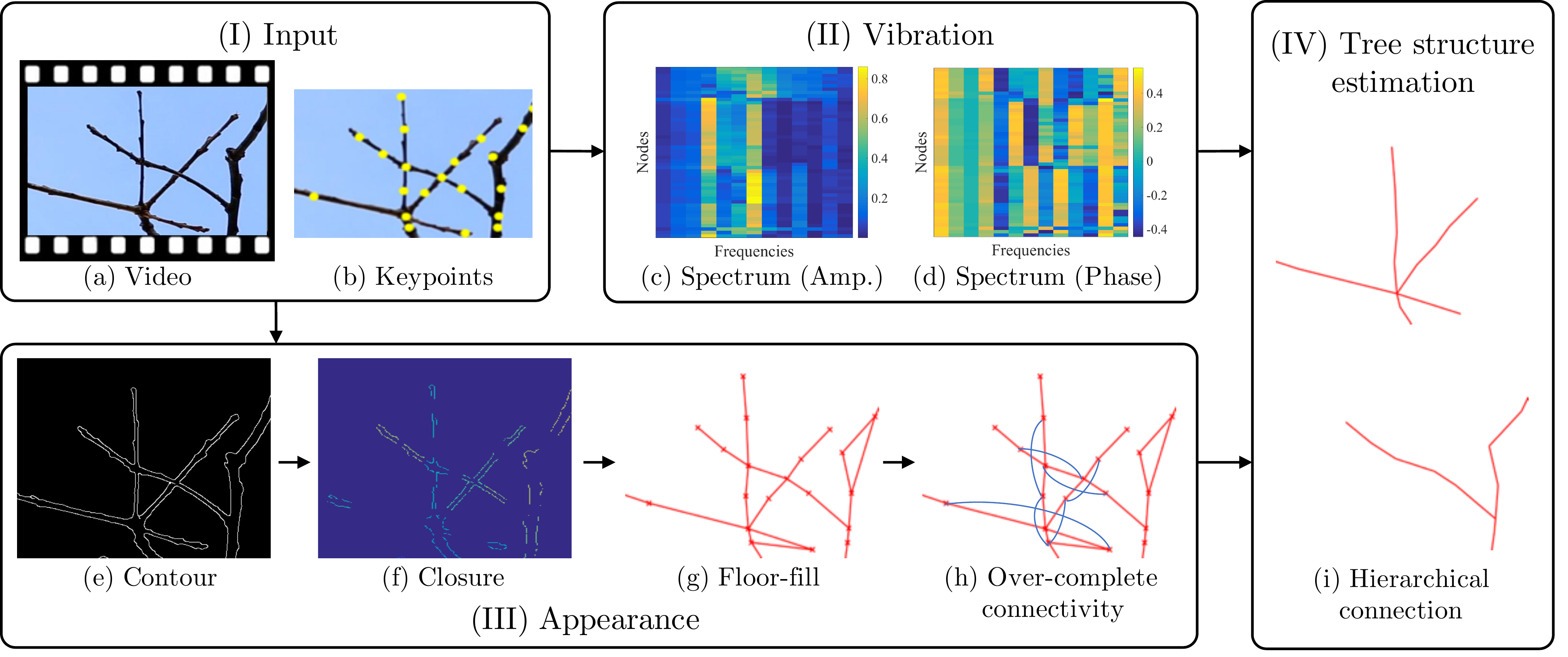}
\vspace{-15pt}
\caption{Overview of our framework. We take a video (a) and a set of keypoints (b) as input (I). We use normalized amplitudes (c) and phases (d) of keypoints as our vibration signals (II); we also obtain appearance cues (III) through several intermediate steps (\sect{sec:recognition}). Finally, we apply our inference algorithm (\sect{sec:inference}) for tree structure estimation.}
\label{fig:pipeline}
\vspace{-20pt}
\end{figure*}

%% file: figText/algorithm.tex
\begin{algorithm}[t!]
  
\SetAlgoLined
\LinesNumbered
\DontPrintSemicolon

{\bf Algorithm} $cluster(\bm{Y}, r)$

\KwData{Nodes with complex spectra $\bm{Y}=\{Y_i\}$ and the root's index $r$} 

\vspace{5pt}

Calculate the free vibration of each node in this tree\;
\For{each node $i$}{
    $Y_i \leftarrow Y_i \; ./ \; Y_{r}$
}
Cluster nodes based on their appearance and frequency\;
Let $\{S_j\}_{j=1,\cdots,k}$ be all $k$ clusters\;
\For{$j = 1, \cdots, k$}{
    Select subroot $r_j$\;
    Call $cluster(\bm{Y}_{S_j}, r_j)$ recursively\;
}

\caption{Our hierarchical clustering algorithm}
\vspace{-5pt}
\label{alg:1}
\end{algorithm}

%% file: figText/cluster.tex
\begin{figure}[t]
\centering
\vspace{-10pt}
\begin{subfigure}[b]{0.25\linewidth}
\includegraphics[width=\linewidth]{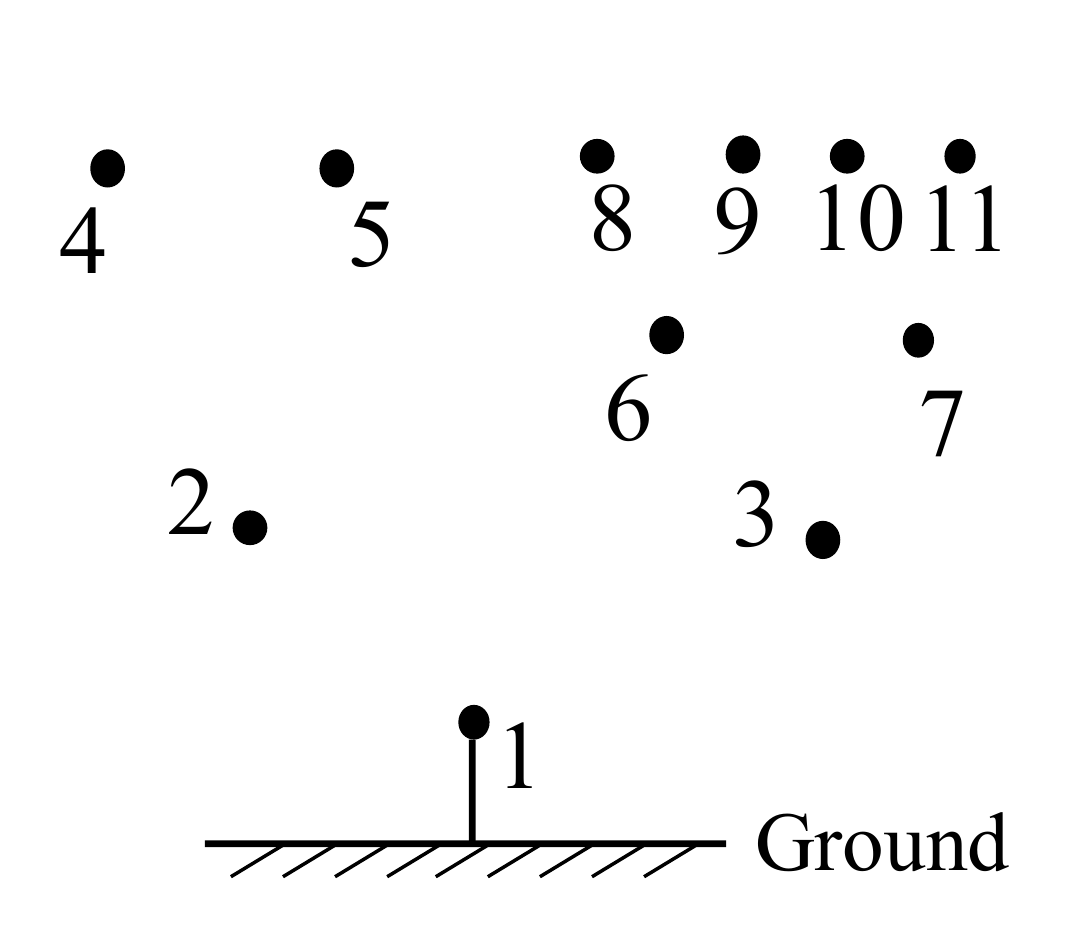}
\caption{Depth 0}
\label{fig:cluster_a}
\end{subfigure}
\quad
\begin{subfigure}[b]{0.25\linewidth}
\includegraphics[width=\linewidth]{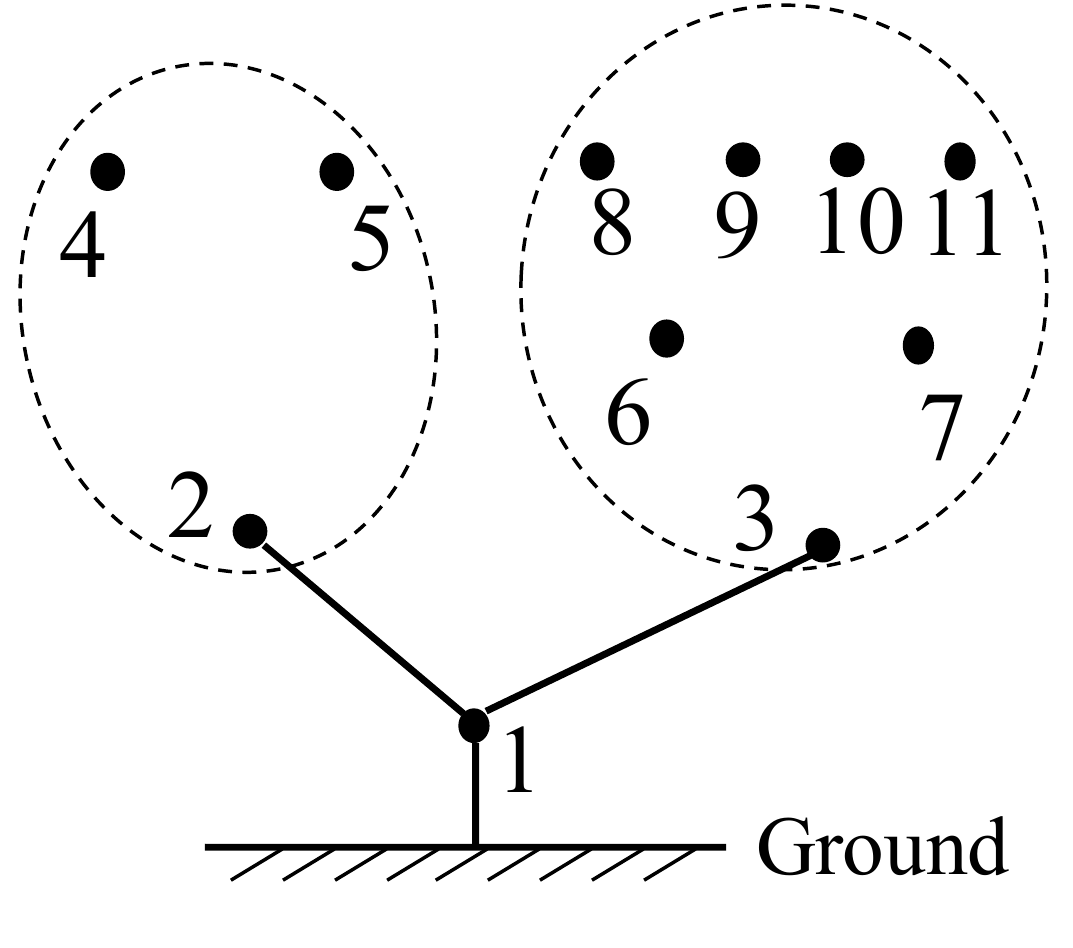}
\caption{Depth 1}
\label{fig:cluster_b}
\end{subfigure}
\quad
\begin{subfigure}[b]{0.25\linewidth}
\includegraphics[width=\linewidth]{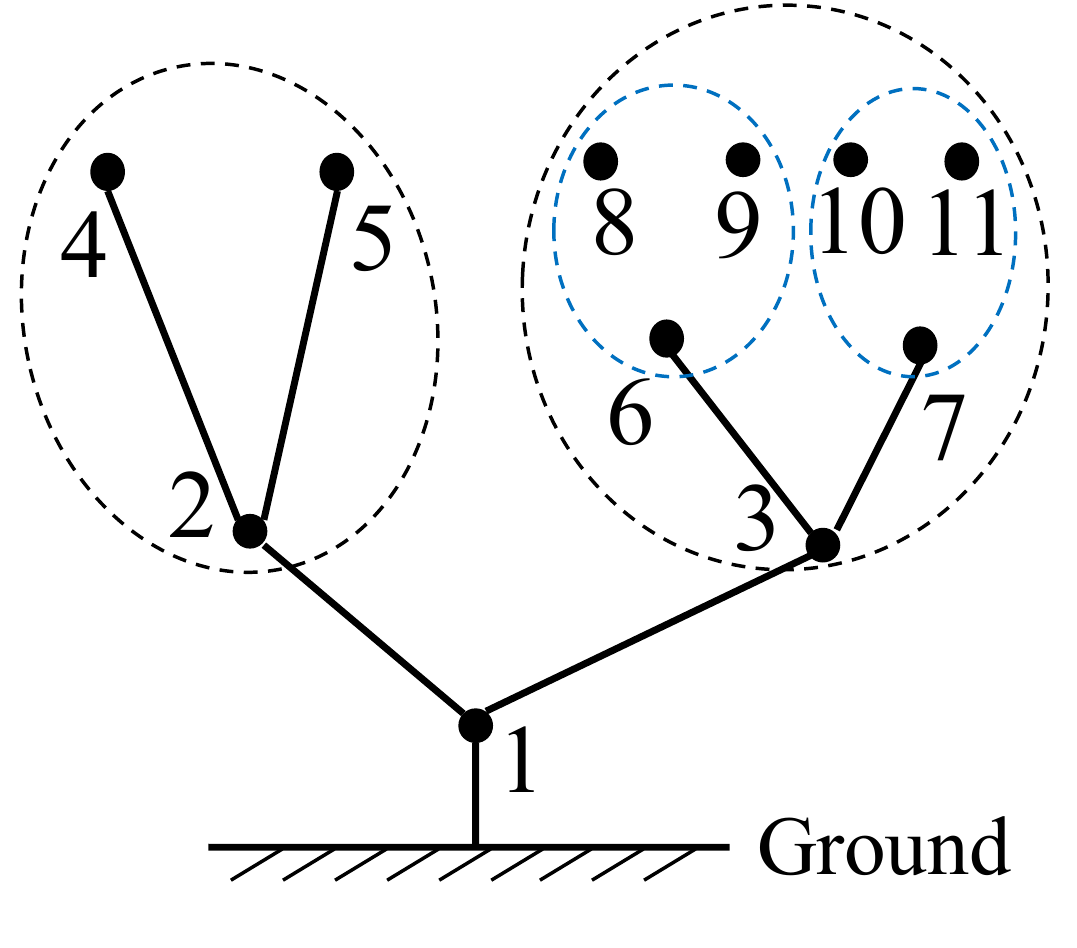}
\caption{Depth 2}
\label{fig:cluster_c}
\end{subfigure}
\vspace{-5pt}
\caption{Illustration of our hierarchical clustering algorithm. See \sect{sec:inference} for details.}
\vspace{-25pt}
\label{fig:h_cluster}
\end{figure}

%% file: text/evaluation.tex
\section{Evaluations}
\label{sec:exp}

We now present how we use simulation to verify our formulation (\sect{sec:formulation}), and show qualitative and quantitative results on videos of artificial and real trees.

\subsection{Simulation}
\label{sec:simulation}

\input{figText/mode}

Based on formulation described in~\sect{sec:problem}, we implemented a tree simulator by solving \eqn{eqn:nl_ode} using the Euler Method~\cite{farlow1993partial}. As shown in \sect{sec:recognition}, the analytic form of ODE is very complicated. Therefore, we do not eliminate all the redundant variables, including the acceleration of the branch ($\ai$ and $\aio$), forces between branches ($\mathbf{r}_i$), and angular velocity of each branch ($\omega_i$). Instead, we directly solve \eqns{eqn:acc_vec} and \ref{eqn:rot_vec} numerically. Also, to increase the stability of Euler method in presence of numerical error, we force the system to have constant total energy for every time-stepping update. If the system's energy increases during an update, we rescale the kinetic and potential energy of each branch to ensure that the total energy of the system is constant. This makes our simulation robust and stable. See the supplementary material for the detailed derivation.

\fig{fig:modes} shows the vibration modes of a simulated tree (left) with three mode shapes (right). Here we manually specify the structure of the tree and physical property of each branch, including mass, stiffness, and length, and numerically solve for the rotation angle of each branch. The mode of power spectra (the natural frequencies) of the trunk and two branches matches the three mode shapes of the tree, which is consistent with the theory in Section~\ref{sec:formulation}.

\subsection{Real, Normal Speed Videos}
\label{sec:real_videos}

\myfirstparagraph{Data. } 
We record videos of both artificial and real trees. For artificial trees, we take 3 videos in an indoor lab environment, where wind is generated by a fan. We take 8 videos of outdoor real trees. All videos are taken at 24 frames per second by a Canon EOS 6D DSLR camera, with a resolution of 1920$\times$1080. 

\myparagraph{Methods. }
We compare our full model, which makes use of appearance and vibration cues jointly (appearance + motion), with a simplified variant, which uses only appearance information, but ignores all motion signals during inference. We also compare with three alternative approaches for hierarchical structure recovery from spatial-temporal motion signals.
\begin{itemize}
    \vspace{-5pt} \item {\bf Appearance + Flow/Tracking:} We replace the spatial-temporal feature in our algorithm by motion recovered by either optical flow or a KLT tracker. 
    \item {\bf Hierarchical motion segmentation:} We use the popular hierarchical video segmentation algorithm~\cite{grundmann2010efficient} to obtain image segments and their structure. We then derive the tree structure from the segment hierarchy. \vspace{-6pt}
\end{itemize}

\input{figText/exp}

\input{figText/realvideo}

\myparagraph{Results. } 
\fig{fig:res} shows that our algorithm works well on real videos. Results in the bottom row suggest that our algorithm can deal with challenging cases. Using motion signals, it correctly recovers the structure of occluded twigs, which is indistinguishable from pure visual appearance.

For quantitative evaluations, we manually label the parents of each node and use it as ground truth. We use two metrics. In \tbl{tbl:res}, we evaluate different methods in (a) the percentage of nodes whose parents are correctly recovered and (b) minimum edit distance---the minimum edges that need to be displaced to make the predicted tree and the ground truth identical. Our algorithm achieves good performance in general. Including motion cues consistently improves the accuracy of the inference on videos of all types, and spatial feature significantly out-performs the raw motion signal.

\subsection{Real, High-Speed Videos}

\myfirstparagraph{Experimental Setup. }
To understand and analyze motion, we take high-speed videos of trees using an Edgertronic high-speed camera. We captured 1 normal-speed video (30 FPS) and 5 high-speed videos with a frame rate varying from 60 to 500 FPS, each of which contains 1,000 frames. For each video, we manually label around 100 interest points and their connections. Intuitively, the root branches should have higher stiffness and lower natural frequencies. Therefore, low-frame-rate videos should provide more information about the tree's main structure, whose natural frequency is low, and high-frame-rate videos should provide more information of fast vibrating thin structure. 

\myparagraph{Evaluation.} For evaluation, we first pick two points ($P_1$ and $P_2$ in \fig{fig:highspeed}c) on two major branches of the tree and compare their power spectra as shown in \fig{fig:highspeed}a. At 60 FPS, the power spectra of these two nodes are different for a wide range of frequencies; at 500 FPS, they are only different at lower frequencies, as the natural frequencies of the main branches are low. We then pick two points ($P_3$ and $P_4$ in \fig{fig:highspeed}c) on two small branches of the tree and compare their power spectra (see \fig{fig:highspeed}b). Now in both 60 FPS and 200 FPS videos, their spectra are similar, and the difference in modes only become significant at 500 FPS. \fig{fig:highspeed}c shows that the estimation errors from low-frame videos (60 or 100 FPS) on the top-right corner no longer exist when the input is at 500 FPS, indicating high-speed videos are better for estimating fine structure. All these results are consistent with our theory. H1 to H6 in \tbls{tbl:res} refer to videos captured at 30, 60, 100, 200, 400, 500 FPS, respectively.

\input{figText/highspeed}

%% file: figText/mode.tex
\begin{figure}[t]
\centering
\includegraphics[width=\linewidth]{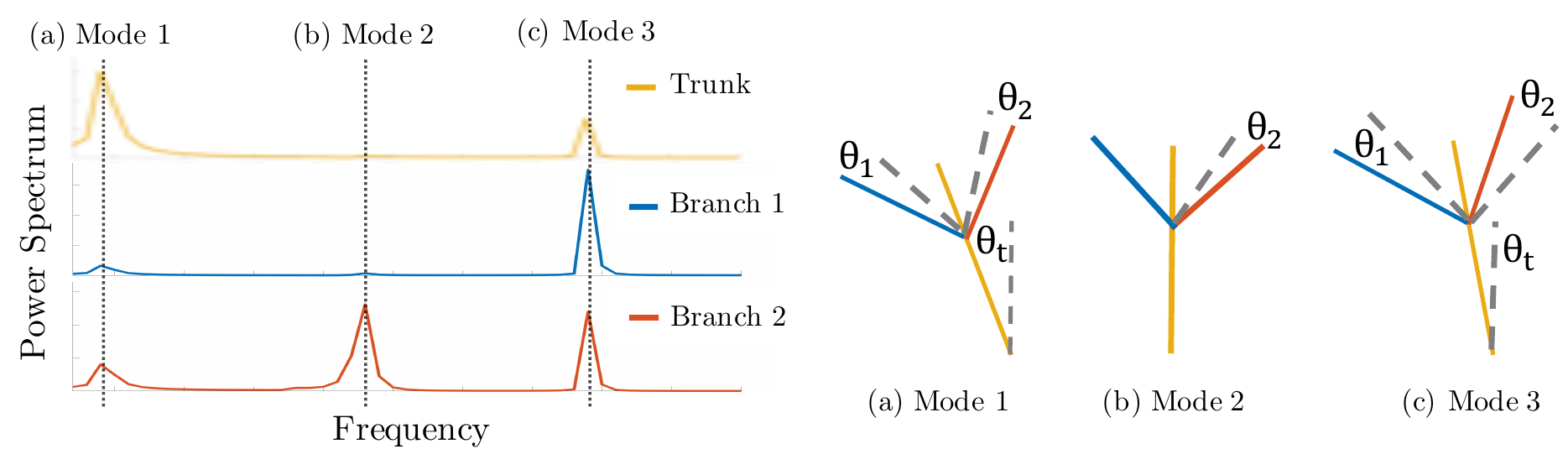}
\vspace{-25pt}
\caption{Mode shapes. The left three curves show the power spectra of the trunk and the two branches. The three mode shapes extracted from vibration are shown on the right.}
 \vspace{-20pt}
\label{fig:modes}
\end{figure}

%% file: figText/exp.tex
\begin{figure*}[t]
\centering
\textbf{}\includegraphics[width=\textwidth]{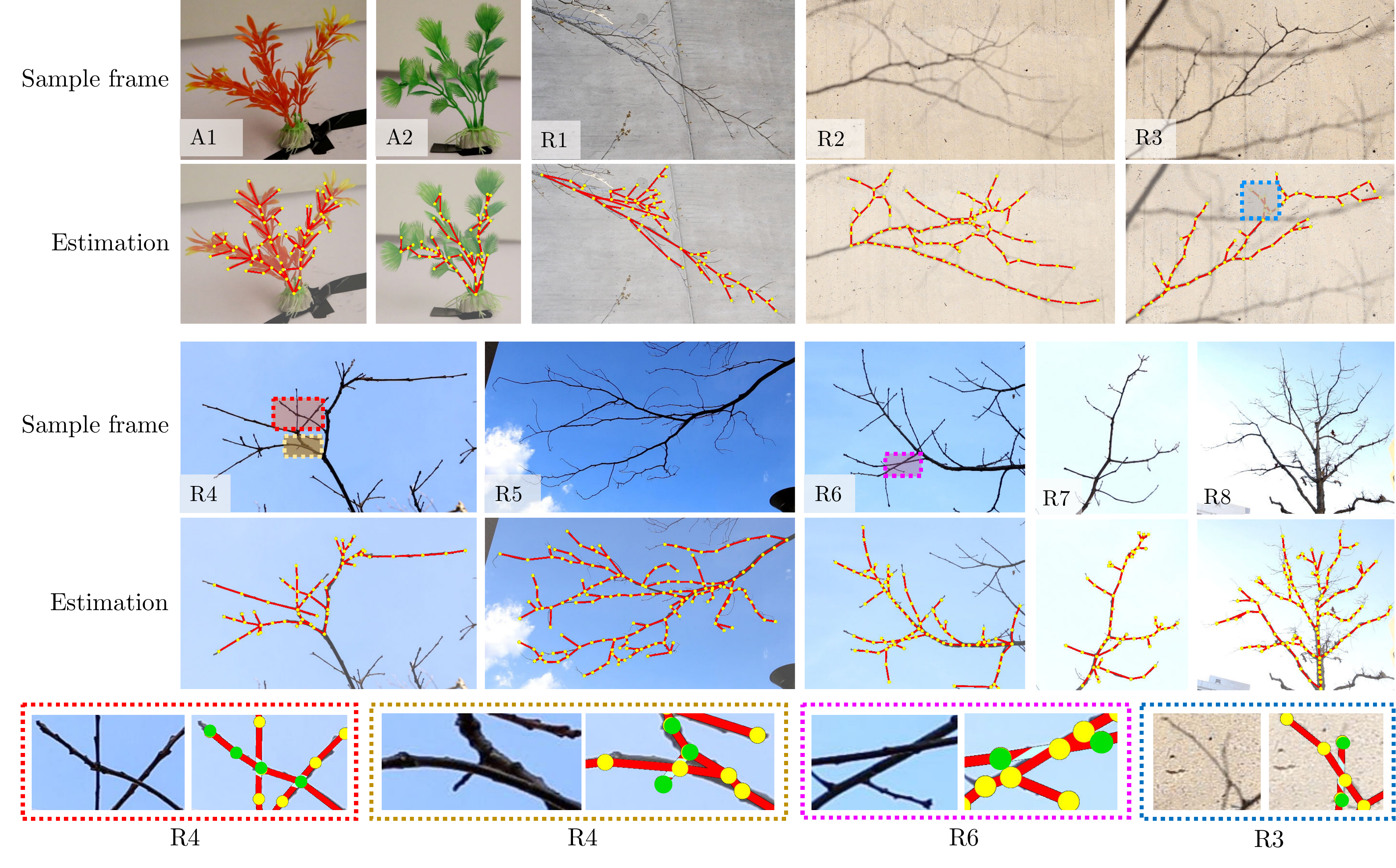}
\vspace{-25pt}
\caption{Estimated tree structure on real videos. A1--A2: on artificial trees; R1--R8: on real trees. At bottom, we show cases where appearance is insufficient for inferring the correct structure. Using vibration signals, our algorithm works well in these cases.}
\vspace{-12pt}
\label{fig:res}
\end{figure*}

%% file: figText/realvideo.tex
\begin{table}[t!]
   \scriptsize
   \centering
    \setlength{\tabcolsep}{2.5pt}
    \begin{tabular}{llcccccccccccccccccc}
    \toprule
    \multirow{2}{*}{Metrics} & \multirow{2}{*}{Methods} & \multicolumn{3}{c}{Artificial } & \multicolumn{8}{c}{Real trees} & \multicolumn{6}{c}{High-speed videos} & \multirow{2}{*}{Avg.} \\
    \cmidrule(lr){3-5}\cmidrule(lr){6-13} \cmidrule(lr){14-19}
    & & A1 & A2 & A3 & R1 & R2 & R3 & R4 & R5 & R6 & R7 & R8 & H1 & H2 & H3 & H4 & H5 & H6 \\
    \midrule
    \multirow{5}{*}{Acc. (\%)} & MoSeg & 33 & 37 & 73 & 50 & 65 & 84 & 56 & 56 & 68 & 70 & 74 & 47 & 57 & 46 & 47 & 43 & 51 & 56.3 \\
    & Appear. & 40 & 31 & 90 & 67 & 59 & 83 & 70 & 66 & 71 & 89 & 85 & 55 & 56 & 62 & 66 & 61 & 69 & 65.7\\
    & A+Flow & 43 & 32 & 92 & 79 & 69 & 83 & 85 & 75 & 84 & 95 & \textbf{94} & 64 & 58 & 69 & 69 & 72 & \textbf{76} & 73.5 \\
    & A+Track & 38 & \textbf{46} & 88 & 79 & 63 & 83 & 84 & \textbf{83} & \textbf{88} & 89 & 93 & 67 & 64 & 66 & 76 & 71 & \textbf{76} & 73.8 \\
    & Ours      & \textbf{54} & 45 & \textbf{100} & \textbf{81} & \textbf{76} & \textbf{94} & \textbf{95} & \textbf{83} & \textbf{88} & \textbf{97} & \textbf{94} & \textbf{69} & \textbf{69} & \textbf{72} & \textbf{77} & \textbf{74} & 70 & \textbf{79.3}\\
    \midrule
    \multirow{5}{*}{Edit Dis.} & MoSeg & 26 & 16 & 7 & 25 & 22 & 8 & 16 & 20 & 13 & 8 & 15 & 20 & 21 & 24 & 22 & 17 & 15 & 17.4 \\
    & Appear. & 20 & 21 & 3 & 12 & 19 & 5 & 5 & 30 & 9 & 4 & 8 & 16 & 16 & 16 & 16 & 16 & 16 & 13.7\\
    & A+Flow & 19 & 13 & 2 & \textbf{7} & 13 & 5 & 3 & 12 & 11 & \textbf{1} & \textbf{6} & 13 & 18 & 11 & 12 & \textbf{8} & \textbf{8} & 9.5\\
    & A+Track & 24 & \textbf{10} & 2 & 10 & 16 & 6 & 4 & 12 & 8 & 4 & 7 & 13 & 15 & 12 & 10 & 10 & \textbf{8} & 10.1 \\
    & Ours & {\bf 14} & 12 & {\bf 0} & 8 & {\bf 12} & {\bf 2} & {\bf 0} & {\bf 6} & {\bf 4} & {\bf 1} & {\bf 6} & {\bf 10} & {\bf 12} & {\bf 9} & {\bf 6} & 9 & {\bf 8} & {\bf 7.0} \\
    \bottomrule
    \end{tabular}
    \vspace{2pt}
    \normalsize
    \caption{Results evaluated by the percentage of nodes whose parents are correctly recovered (top) and the edit distance between reconstruction and ground truth (bottom). Our method outperforms the alternatives in most cases.}
    \label{tbl:res}
    \vspace{-28pt}
\end{table}

%% file: figText/highspeed.tex
\begin{figure*}[t]
\centering
\includegraphics[width=\textwidth]{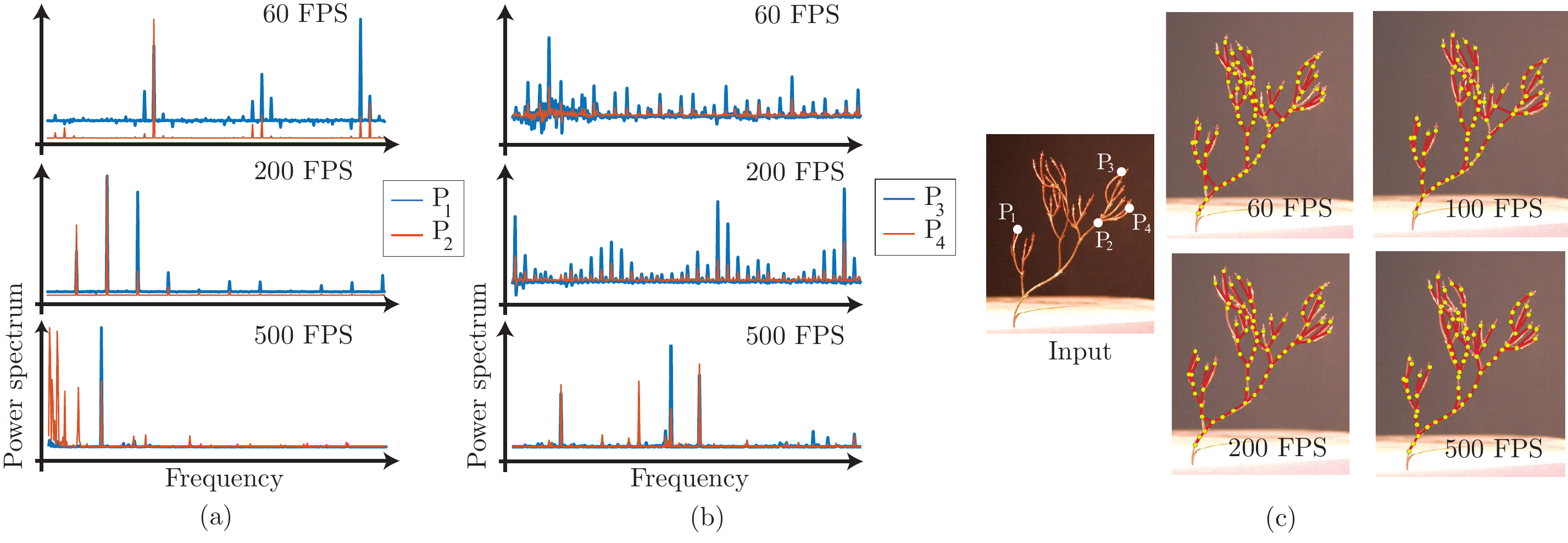}
\vspace{-20pt}
\caption{Evaluation of the algorithm on videos with different frame rates. (a) and (b) shows the power spectra of selected nodes in the input videos captured at different frame rates, and (c) shows the estimated tree structures. See \sect{sec:real_videos} for more details.}
\label{fig:highspeed}
\vspace{-20pt}
\end{figure*}

%% file: text/application.tex
\section{Applications}

Our model has wide applications in inferring tree-shaped structure in real-life scenarios. To demonstrate this, we show two applications: seeing object structure from shadows, and inferring blood vessels from retinal videos.

\myparagraph{Shapes from Shadows. }
In circumstances like video surveillance, often the only available data is videos of projections of an object, but not the object itself. For example, we can see the shadows of trees in the video, but not the trees themselves. In these cases, it would be of strong interests to reconstruct the actual shape of the object. Our algorithm deals with these cases well. Among the eight real videos in \fig{fig:res}, R2 and R3 are videos of tree shadows. Our algorithm successfully reconstructs the underlying tree structure, as shown in \fig{fig:res} and \tbl{tbl:res}.

\input{figText/retinal}

\myparagraph{Vessels from Retinal Videos. }
Our model can contribute to biomedical research. We apply our model on a retinal video from OcuScience LLC. As shown in \fig{fig:retinal}a-b, our algorithm performs well, reconstructing the connection among retinal vessels despite limited video quality. It achieves a smaller edit distance (4) compared with A+Flow (7) and A+Track (6).

\myparagraph{Fully Automatic Recovery. }
While we choose to take keypoints as input to provide users with extra flexibility and to increase prediction accuracy, following the convention in the literature~\cite{turetken2016reconstructing}, our system can be easily extended to become fully automatic. Here we provide an additional experiment on the retinal video. We first apply the segmentation method from Mannis~\etal~\cite{maninis2016deep} on the first frame to obtain a segmentation of vessels. We then employ the classical skeletonization algorithm from Lee~\etal~\cite{lee1994building} (\fig{fig:retinal}c), and use the endpoints and junctions of the obtained skeleton as input keypoints to our model. As shown in \fig{fig:retinal}d, our system works well without manual labels.

%% file: figText/retinal.tex
\begin{figure}[t]
\centering
\includegraphics[width=\linewidth]{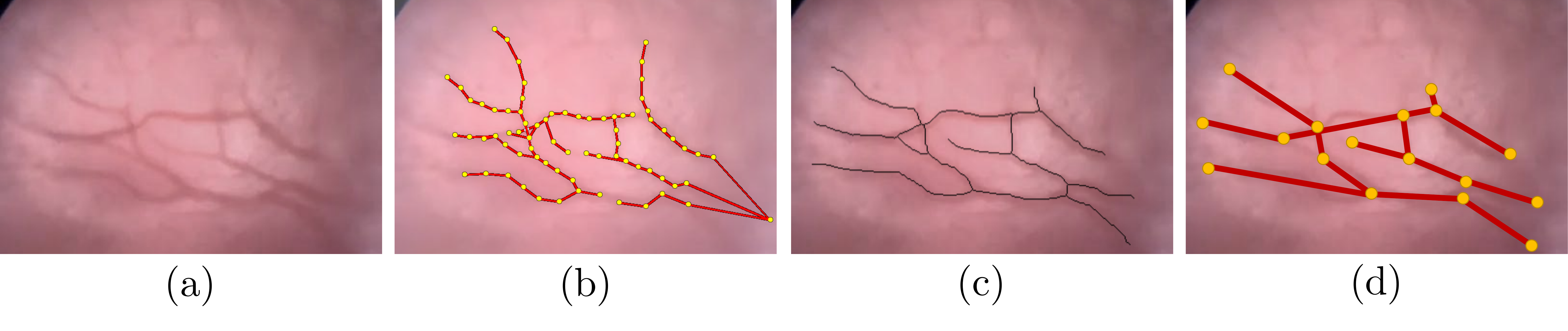}
\vspace{-23pt}
\caption{Our result on a retinal video. (a) A frame from the input video. (b) Our model reconstructs the structure of blood vessels despite low video quality. (c-d) Results on fully automatic structure inference, where (c) shows the estimated object skeleton and (d) shows the object structure inferred by our model.}
\label{fig:retinal}
\vspace{-25pt}
\end{figure}

%% file: text/discussion.tex
\section{Discussion}

In this paper, we have demonstrated that vibration signals in the spectral domain, in addition to appearance cues, can help to resolve the ambiguity in tree structure estimation. We designed a novel formulation of trees from physics-based link models, from which we distilled physical properties of vibration signals, and verified them both theoretically and experimentally. We also proposed a hierarchical inference algorithm, using nonparametric Bayesian methods to infer tree structure. The algorithm works well on real-world videos.

Our derivation makes four assumptions: passive motion, small vibration, no damping, and a known root. While real trees often satisfy the first two, they do not have zero damping (damping ratio ranging from 1.2\% to 15.4\%~\cite{james2014branches}). In these cases, our algorithm still successfully recovers their geometry from vibration. When the root is unknown, our method can discover multiple subtrees from a virtual root with a uniform motion spectrum. On the other hand, our model performs less well when assumptions are significantly violated (\eg, large vibration or an incorrect root). 

We see our work as an initial exploration on how spectral knowledge may help structured inference, and look forward to its potential applications in fields even outside computer science, \eg, fiber structure estimation.

%% file: text/supp.tex
\begin{figure}[t]
\centering
\includegraphics[width=0.6\linewidth]{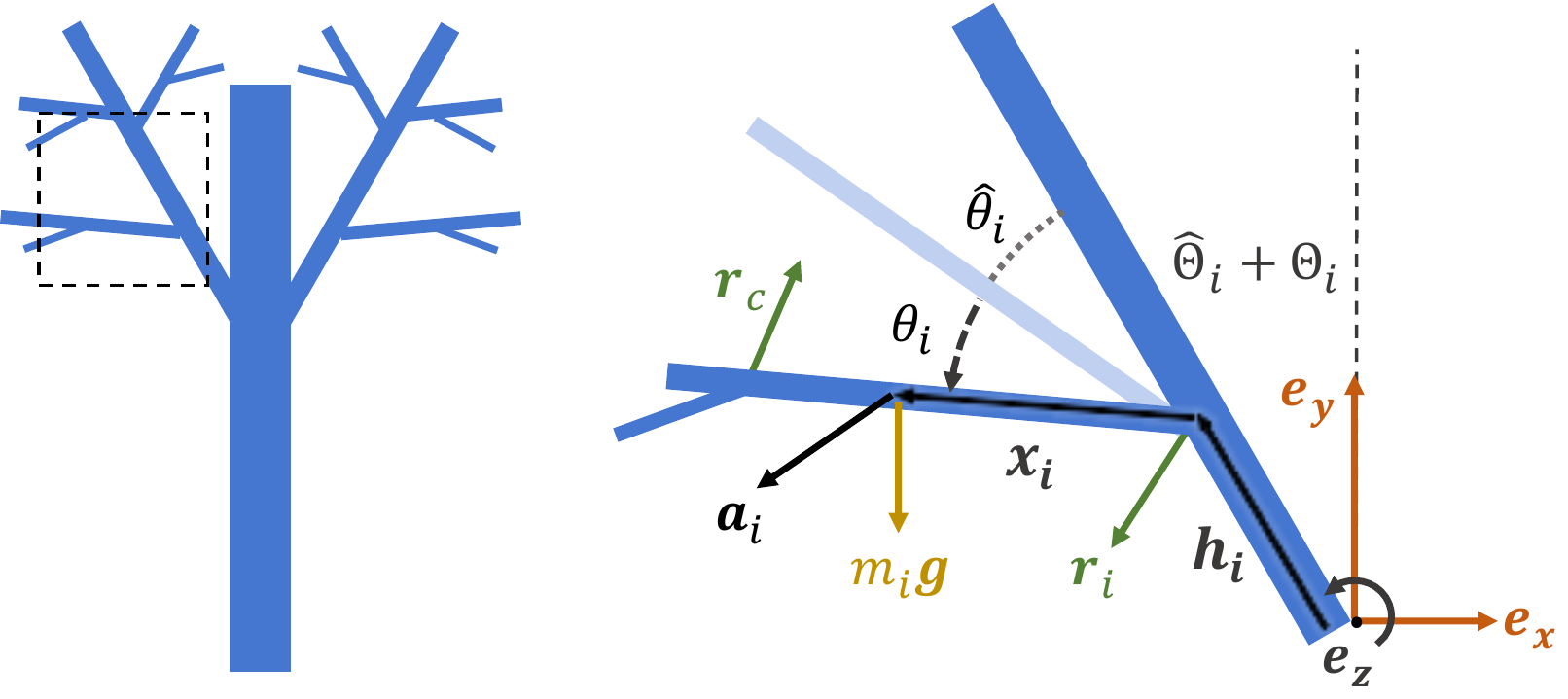}
\caption{Hierarchical beam structure.}
\label{fig:coord}
\end{figure}

\section{Derivation of ODE}
As discussed in Section 3.1 in the main paper, each branch $i$ has to satisfy the following 5 equations:
\begin{align}
    m_i \ai & = -\ri + \sum_{c \in C_i} \rc + m_i \g, \label{eqn:acc_vec} \\
    I_i \dot{\omega}_i & = -k_i \htheta_i + \sum_{c\in C_i} k_c \htheta_c + \ri \times \x_i + \sum_{c\in C_i}\rc \times \x_i,\label{eqn:rot_vec} \\
    \ai & = \aio + \dot{\omega}_i \times \x_i + \omega_i \times (\omega_i \times \x_i), \label{eqn:acc_constraint_vec}\\
    \omega_i & = \dhtheta_i + \sum_{p\in P_i}\dhtheta_p, \label{eqn:agvlc}\\
    \dot{\omega}_i & = \ddhtheta_i + \sum_{p\in P_i}\ddhtheta_p, \label{eqn:agacc}
\end{align}
where all vectors are defined in a fixed global coordinate system, as shown in \fig{fig:coord}. The unknowns are $\ai$, $\aio$, $\ri$, $\omega_i$,  $\dot{\omega}_i$, $\htheta_i$, $\dhtheta_i$, $\ddhtheta_i$. $\ai$ represents the acceleration of branch $i$ at its center of mass. $\aio$ is the acceleration of branch $i$'s junction with its parent. $\ri$ is the force branch $i$ applied to its parent.  $\omega_i$, $\dot{\omega}_i$ are branch $i$'s angular velocity and angular acceleration respectively. $\htheta_i$ is the deviation angle of  branch $i$ from the equilibrium position. $\dhtheta_i$ and $\ddhtheta_i$ are the first and second order derivatives of $\htheta$ with respect to time, respectively.  As we are only considering the 2D motion of the tree, all rotation vectors (angular velocity $\omega$ and angular acceleration $\dot\omega$) in \eqn{eqn:rot_vec} are perpendicular to the plane of the tree (image plane), and align with the direction of $\bm r\times \x$. With Equations~(\ref{eqn:acc_vec})-(\ref{eqn:agacc}), we aim to derive the ODE in the form of Equation (5) in Section 3.1 of the main paper
\begin{equation}
     g(\htheta, \dhtheta, \ddhtheta)=0. \label{eqn:nl_ode}
\end{equation}

To start with, we substitute all $\omega_i$ and $\dot{\omega}_i$ with $\dhtheta_i$ and $\ddhtheta_i$, respectively, using \eqns{eqn:agvlc} and (\ref{eqn:agacc}). 
For the ease of description, we define
\begin{equation}
    f_{i}(\theta) \defeq \theta_i+\sum_{p\in P_i} \theta_p,
\end{equation}
from which we have
\begin{equation}
    \omega_i = f_i(\dhtheta),\quad \dot{\omega}_i = f_i(\ddhtheta) \label{eqn:domegai}.
\end{equation}

Then, one could recursively substitute $\ai$ using
\begin{equation}
    \aio  = \apo + f_{p}(\ddhtheta) \times \h_i +  f_{p}(\dhtheta)  \times (f_{p}(\dhtheta) \times \h_i), \label{eqn:juntion_acc}
\end{equation}
where $\h_i$ is the vector that starts from its parent's junction and points to its own junction point. As the junction of the trunk has zero acceleration, all $\aio$ and $\ai$ can be replaced by a function of $\htheta$, $\dhtheta$, and $\ddhtheta$, note as\footnote{With the abuse of notation, here $\ai$ denotes a 2D vector of the acceleration of branch i, and $\ai(\htheta,\dhtheta,\ddhtheta)$ denotes a function that maps $\htheta$, $\dhtheta$, and $\ddhtheta$ to the acceleration.}
\begin{equation}
 \fa{i} \defeq \ai \label{eqn:ai}.
\end{equation}
To substitute all $\ri$, one could start with any leaf branch $l$, so that \eqn{eqn:acc_vec} becomes
\begin{equation}
    m_i \fa{l} = -\rl + m_i \g.
\end{equation}
For all the leaf branches, we define
\begin{equation}
\fr{l} \defeq - m_i \fa{l} + m_i \g,
\end{equation}
so we can write $\rl$ as $\fr{l}$.

Then we iteratively substitute all $\ri$ from leaf branches to the trunk, \ie,
\begin{equation}
    m_i \fa{i} = -\ri + \sum_{c \in C_i} \fr{c} + m_i \g, 
\end{equation}
and we define
\begin{equation}
\fr{i} = \sum_{c \in C_i} \fr{c} + m_i \g - m_i \fa{i}, \label{eqn:ri}
\end{equation}
for all non-leaf branches. Then $\ri$ is a function $\fr{i}$ of $\htheta$, $\dhtheta$, and $\ddhtheta$ for all the branches.

Substituting \eqn{eqn:rot_vec} with \eqns{eqn:domegai} and (\ref{eqn:ri}), we have
\begin{equation}
    I_i f_{i}(\ddhtheta)  = -k_i \htheta_i + \sum_{c\in C_i} k_c \htheta_c + \fr{i} \times \x_i + \sum_{c\in C_i}\fr{c} \times \x_i, 
\end{equation}
which is the desired ODE system of the form
\begin{equation}
     g(\htheta, \dhtheta, \ddhtheta)=0. \nonumber
\end{equation}

\section{Linearization of ODE}
In this section, we show how to linearize \eqn{eqn:nl_ode} under the assumption that $\htheta$ is small. As mentioned in the main text, we assume small vibrations, \ie, $O(\htheta^2)$, $O(\dhtheta^2)$, and $O(\ddhtheta^2)$ are negligible. 

Based on this assumption, we investigate all the non-linear terms in \eqns{eqn:acc_vec} to (\ref{eqn:agacc}). Note that only \eqns{eqn:rot_vec} and (\ref{eqn:acc_constraint_vec}) contain non-linear terms. Specifically, for \eqn{eqn:rot_vec}, we have (using the scalar form here for clarity)
\begin{align}
     I_i \dot{\omega}_i  = &-k_i \htheta_i + \sum_{c\in C_i} k_c \htheta_c+ (r_{ix}+r_{cx})\frac{l_i}{2}\cos\left(\htheta_i + \hat{\theta}_i + \Theta_i + \hat{\Theta}_i\right), \nonumber \\
     & + (r_{iy}+r_{cy})\frac{l_i}{2}\sin\left(\htheta_i + \hat{\theta}_i + \Theta_i + \hat{\Theta}_i\right)
     ,\label{eqn:rot_scalar}
\end{align}
where
\begin{align*}
     \Theta_i &= \sum_{p\in P_i} \theta_p, &\hat{\Theta}_i& = \sum_{p\in P_i} \hat{\theta}_p,  \\
     \ri &= r_{ix}\mathbf{e_x} +r_{iy}\mathbf{e_y}, &\rc& = r_{cx}\mathbf{e_x} +r_{cy}\mathbf{e_y}.
\end{align*}
Under the assumption that $\theta$ is small, we linearize the $\sin(\cdot)$ and $\cos(\cdot)$ terms through the first-order Taylor expansion (as both ${\Theta}_i$ and ${\theta}_i$ are very small, their higher-order terms are negligible),
\begin{align*}
     \sin\left(\htheta_i + \hat{\theta}_i + \Theta_i + \hat{\Theta}_i\right) \approx \sin\left(\hat{\Theta}_i + \hat{\theta}_i\right)+\cos\left(\hat{\Theta}_i + \hat{\theta}_i\right)({\Theta}_i + {\theta}_i), \\
     \cos\left(\htheta_i + \hat{\theta}_i + \Theta_i + \hat{\Theta}_i\right) \approx \cos\left(\hat{\Theta}_i + \hat{\theta}_i\right)-\sin\left(\hat{\Theta}_i + \hat{\theta}_i\right)({\Theta}_i + {\theta}_i).
\end{align*}

For \eqn{eqn:acc_constraint_vec}, we have
\begin{align}
        a_{ix} &= a_{i_ox} + \dot{\omega}_i\left[-\frac{l_i}{2}\cos\left(\htheta_i + \hat{\theta}_i + \Theta_i + \hat{\Theta}_i\right)\right] + \omega_i^2\left[\frac{l_i}{2}\sin\left(\htheta_i + \hat{\theta}_i + \Theta_i + \hat{\Theta}_i\right)\right], \label{eqn:aix_exp}\\
        a_{iy} &=  a_{i_oy} +  \dot{\omega}_i\left[-\frac{l_i}{2}\sin\left(\htheta_i + \hat{\theta}_i + \Theta_i + \hat{\Theta}_i\right)\right] + \omega_i^2\left[-\frac{l_i}{2}\cos\left(\htheta_i + \hat{\theta}_i + \Theta_i + \hat{\Theta}_i\right)\right] \label{eqn:aiy_exp}.
\end{align}
Taking the first-order Taylor expansion to the equations above, we have
\begin{align}
        a_{ix} &\approx a_{i_ox} + \dot{\omega}_i\left[-\frac{l_i}{2}\cos\left(\hat{\Theta}_i + \hat{\theta}_i\right)\right],\\
        a_{iy} &\approx a_{i_oy} +  \dot{\omega}_i\left[-\frac{l_i}{2}\sin\left(\hat{\Theta}_i + \hat{\theta}_i\right)\right].
\end{align}
where terms of the form $\dot\omega\theta_i$ are omitted as $\dot\omega\theta_i \leq \frac{1}{2}\left(\dot\omega^2+\theta_i^2\right)$ is a second-order term. Also, the last term in \eqns{eqn:aix_exp} and (\ref{eqn:aiy_exp}) are omitted as they contain the second-order factor $\omega_i^2$. Then, using the same elimination techniques from the previous section, we obtain a set of linear equations with respect to $\theta$ and $\ddhtheta$, \ie,
\begin{equation}
     M\ddot{\bhtheta} + K\bhtheta = \bm{0}, \nonumber
\end{equation}
which is Equation (6) in the main manuscript.

\section{Simulation Details}

We built a tree simulator by solving \eqn{eqn:nl_ode} using the Euler method. Let $t$ be the temporal index, and $\theta(t)$ be all the rotations observed at time $t$. The idea is that given the initial condition ($\theta(0)$, $\dhtheta(0)$), we can solve $\ddhtheta(0)$ by inverting a linear system. Then, $\theta(\Delta t)$ and $\dhtheta_{\Delta t}$ are approximated by $\theta(0) + \dhtheta(0)\Delta t$ and $\dhtheta(0) + \ddhtheta(0)\Delta t$, respectively.

Theoretically, one could explicitly write out the linear system by performing the substitution described earlier. However, it is much more convenient to introduce intermediate variables $\ai$, $\aio$, $\ri$, and $\dot{\omega}_i$, and solve the linear system built from \eqns{eqn:acc_vec} to (\ref{eqn:agacc}) and \eqn{eqn:juntion_acc}. By gathering the coefficients at time $t$, one could write all linear equations above as
\begin{align}
    A \begin{bmatrix}
           \ddot{\bhtheta}(t) \\
           \mathbf{r}(t) \\
           \mathbf{a}(t) \\
           \mathbf{a_{o}}(t)\\
           \mathbf{\dot{\omega}}(t)
         \end{bmatrix} = \bm b.
  \end{align}
Then we can solve $\ddot{\bhtheta}(t)$ through the linear system above.